\documentclass[10pt,journal, finalsubmission, a4paper ]{IEEEtran}

\usepackage{amsmath}
\usepackage{tikz, rotating, cases, multirow,scalefnt }
\usepackage{amssymb}
\usepackage{color, array}
\usepackage{mathbbol}

\usepackage{times}

\usetikzlibrary{arrows}
\tikzset{>=stealth'}
\tikzstyle{edge}  =[draw,-] 

\def\secref{Section~\ref}
\def\figref{Figure~\ref}

\def\eqref{Equation~\ref}

\DeclareMathOperator*{\argmax}{arg\,max}
\def\nu#1#2{{#1}\kern.166em\textrm{#2}}

\def\usertext#1{``\emph{#1}''}%
\def\usersym#1{\emph{#1}}%
\def\nu#1#2{{#1}\kern.166em\textrm{#2}}%

\def\sym#1{\text{#1}}%
\def\qsym#1{\text{``\sym{#1}''}}%
\def\qtxt#1{\text{``\text{#1}''}}%

\begin{document}


\title{The Statistical Model for Ticker, an Adaptive Single-Switch Text-Entry Method for Visually Impaired Users}

\author{
\begin{tabular}{cccc}
Emli-Mari Nel &  Per Ola Kristensson & David~J.~C.~MacKay \\
University of Cambridge &    University of Cambridge  & University of Cambridge \\
en256@cam.ac.uk &  pok21@cam.ac.uk & djcm1@cam.ac.uk \\
\end{tabular}
}

\maketitle

\begin{abstract}
  This paper presents the statistical model for Ticker~\cite{ticker2018_tpami}, a novel probabilistic stereophonic single-switch text entry method for visually-impaired users with motor disabilities who rely on 
  single-switch scanning systems to communicate. All terminology and notation are defined in~\cite{ticker2018_tpami}.  

\end{abstract}

{ \IEEEkeywords single-switch systems, accessibility,  augmentative and alternative communication, Bayesian inference.}


\section{Letter selections from audio files}

\begin{figure}[!!!!!htbp]
  \centering
  \begin{tabular}{c}

\begin{minipage}{7.8cm} {\small fqwaglrxbhmsycintzdjou\_ekpv.dimrwejnsxakotybgpuzcflv\_hq}.\end{minipage}     \\ (a) \\ 

\begin{minipage}{7.8cm} \centering  \includegraphics[width=7.8cm]{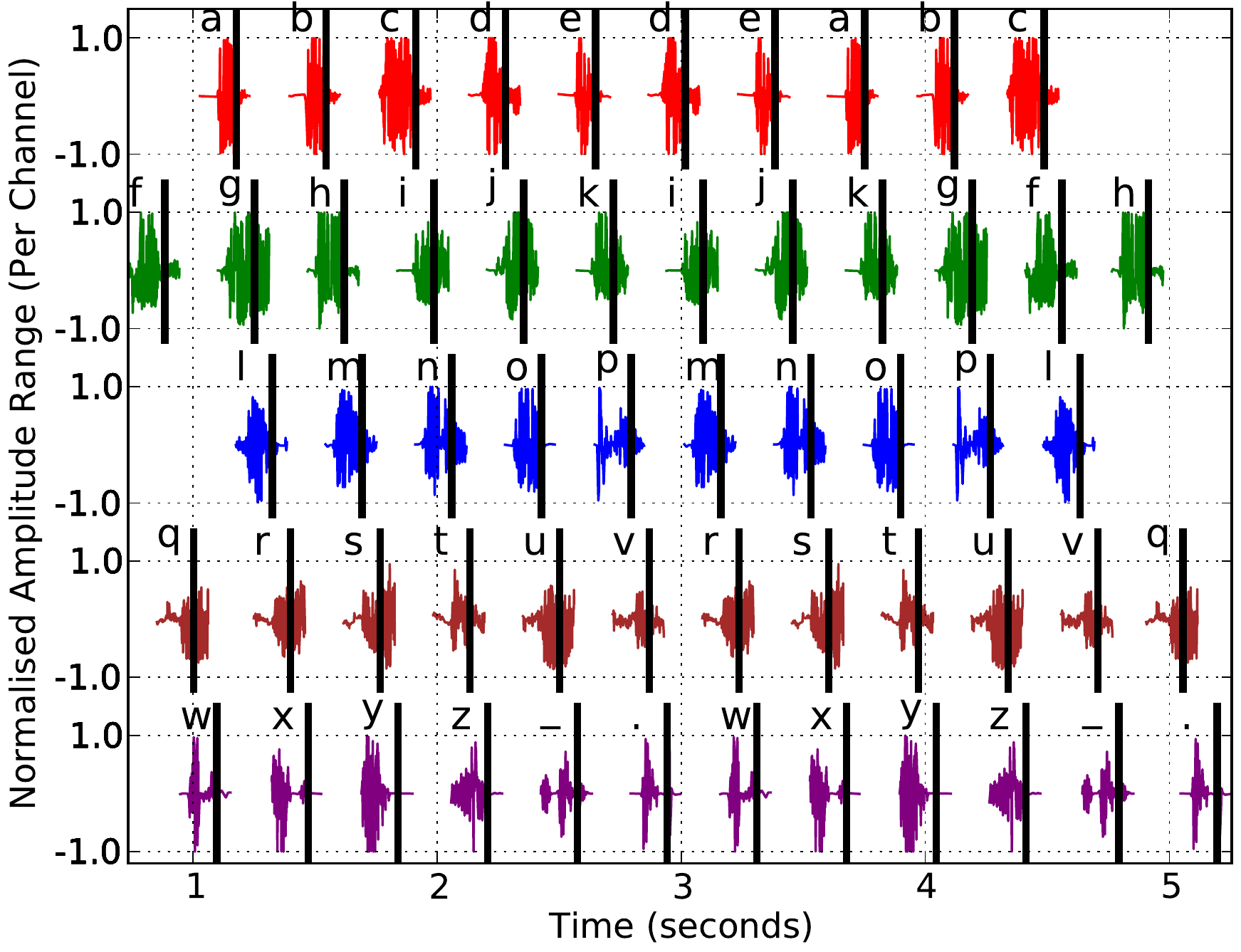} \end{minipage}    \\ (b) \\
 \end{tabular}
  \caption{(a) A composite audio sequence that can be presented to the user for $R=2$. Ideally the user should click twice per letter, but an accidental miss can be tolerated. (b) The normalised amplitudes of all sound files  for Ticker in five channel mode. That is, the composite sequence in (a) is presented to the user, where five different voices read the alphabet to the user. Sounds within the same channel are indicated with the same colour. 
 }
  \label{fig:ampl}
\end{figure}

In  \figref{fig:ampl}(a) a typical {\em composite} audio sequence that can be presented to the user is shown, where the composite sequence consists of two repetitions of the alphabet.  In Ticker, the user selects one letter at a time when listening to such a sequence. In the shown example, the user can click twice per letter. The second repetition occurs in a different order than the first, which allows one to infer the intentional letter selection more accurately.

The system does not explicitly make any selection after a click is received; instead the system accumulates evidence. 
After one or more clicks are received, the system internally updates the posterior word probabilities. 
It will then proceed to play the composite sequence again for the next letter.   
When the posterior probability of any word in a pre-defined dictionary is above a certain threshold, that word is selected. 




We have shown in~\cite{ticker2018_tpami} how to effectively parallelise the audio input stream: Groups of letters are uttered in the same audio channel by the same person, as illustrated in~\figref{fig:ampl}(b). The user is expected to wear headphones. The letter ``a" is, for example, always uttered in the user's left ear by the same voice, whereas the letter ``z" is uttered by a different voice in the user's right ear. If the user is able to focus on a specific voice, the brain will tend to filter everything else out by virtue of the cocktail-party effect. In the shown example, it allows one to play the alphabet twice to the user in just over five seconds. We refer to the main paper \cite{ticker2018_tpami} for an overview of the system and definitions of terminology and notation.


\section{The composite audio sequence for $R=2$}
\label{sec:audio_seq}

In this section we describe how to derive the composite sequence for Ticker in two channel mode. We focus on $R=2$, as this is the default setting for Ticker. $R=1$ is suitable for very noisy switches, or when the user is unable to distinguish between sounds in stereophonic mode. With $R=2$, two clicks per selection is usually necessary, which is directly comparable to standard scanning systems. This setting is intended for situations where the user is conscious and has the capability to 
memorise the composite audio sequence. A typical user would be \emph{compos mentis}, and use, for example, a blink detector to click within a few seconds of when they intend  to.

Due to the serial nature of the interface for this application, it is difficult to come up with a technique that adapts to the interface
dynamically (so that more probable words/letters are easier to select) without increasing the cognitive load too much.  
We therefore assume the composite audio sequence to be fixed so that the user can pre-empt when a certain letter will be pronounced.

We assume the letters occur in alphabetical order   within each clip for $r=1$, so that the user only has to memorise the clip for $r=2$. 
For example, in \figref{fig:ampl}(b), the letter sequence \usersym{abcde}
of the clip associated with the voice shown in red
is associated with $r=1$. The user has to then memorise 
  \usersym{deabc} for $r=2$.

If the user's click-time precision is noisy, it can be difficult
to make an estimate of the intended letter after one click.
If the letters that were close to the intended letter at $r=1$ occur far away from it at $r=2$, disambiguation can become remarkably easier.
Hence, to compute the composite audio sequence for $r=2$, the distances between letters that were close to each other for $r=1$ should be large for $r=2$.
%
Since all sound files are assumed to have the same length, one can integerise this computation,
considering the number of letters between certain letter pairs.
For example, in \figref{fig:ampl}(b), letters \usersym{r} and \usersym{x} are adjacent for $r=1$, whereas they are separated by five letters for $r=2$.

Let $A$ be the number of letters in the alphabet ($\forall r \in \{1, \ldots, R\}$). The total number of letters in the composite sequence is then $AR$, which is not always divisible 
by the number of channels. This can cause the characters to sound
arrhythmic at the beginning and end of  a clip, 
making it more difficult to tune in on a voice.
To account for this, some sound files at the beginning/end of
a clip were made slightly longer.  
To further assist the user to control  his/her timing, two ``tick'' sounds
were added at the beginning of the composite audio sequence to set the pace of the rhythm.
The adjusted  sound-file lengths and the addition of the ``tick'' sounds 
were not part of the interface during the initial user trials. These adjustments were made
after feedback from the users, and resulted in a significant improvement.

The computation of the composite audio sequence is performed in two steps.
Firstly, the $K$ nearest neighbours of each letter $\ell$ are stored for $r=1$.
Secondly, the sequence for $r=2$ is chosen such that all of the stored
neighbours from the first step are at least $K$ letters away from~$\ell$.
This process is repeated to maximise~$K$.
It was found that for 1--5 channels, the maximal $K$ are $(4,4,4,3,3)$, respectively. 

There can be several sequences with the same $K$.
Some of the sequences were further eliminated by restricting
successive sounds, as some sounds can become indistinguishable when they overlap.
Sequences containing any of the following successive letters,
\{\usersym{a},\,\usersym{h}\},
\{\usersym{q},\,\usersym{k}\},
\{\usersym{m},\,\usersym{n}\},
\{\usersym{b},\,\usersym{d}\} and
\{\usersym{a},\,\usersym{i}\} were removed.

\begin{figure}[!!!!!htb]
  \centering      
 
   \begin{tabular}{|c@{}|} \hline   
   
    \begin{tabular}{l@{}}   
   {\hspace{-0.5cm} \small 1)  abcdefghijklmnopqrstuvwxyz\_.wrmhczupkfaxsnid\_vqlgbytoje.} \\
   {\hspace{-0.5cm} \small 2)  aobpcqdresftguhviwjxkylzm\_n.lwgrb\_kvfqazjuepnyitdomxhsc.} \\
   {\hspace{-0.5cm} \small 3)  sajtbkuclvdmwenxfoygpzhq\_ir.fmuaqyelsipxdk.howcj\_gnvbrz} \\
   {\hspace{-0.5cm} \small 4)  ahovbipwcjqxdkryelszfmt\_gnu.bjrzgiqyfnowemuxalp\_dhs.cktv} \\                                               
   {\hspace{-0.5cm} \small 5)  fqwaglrxbhmsycintzdjou\_ekpv.dimrwejnsxakotybgpuzcflv\_hq.} \\
   \end{tabular}  \\
     
     \begin{tabular}{c@{}c@{}c@{}} 
    \hspace{-0.6cm} 
    \begin{minipage}{2.7cm} \includegraphics[width=2.7cm]{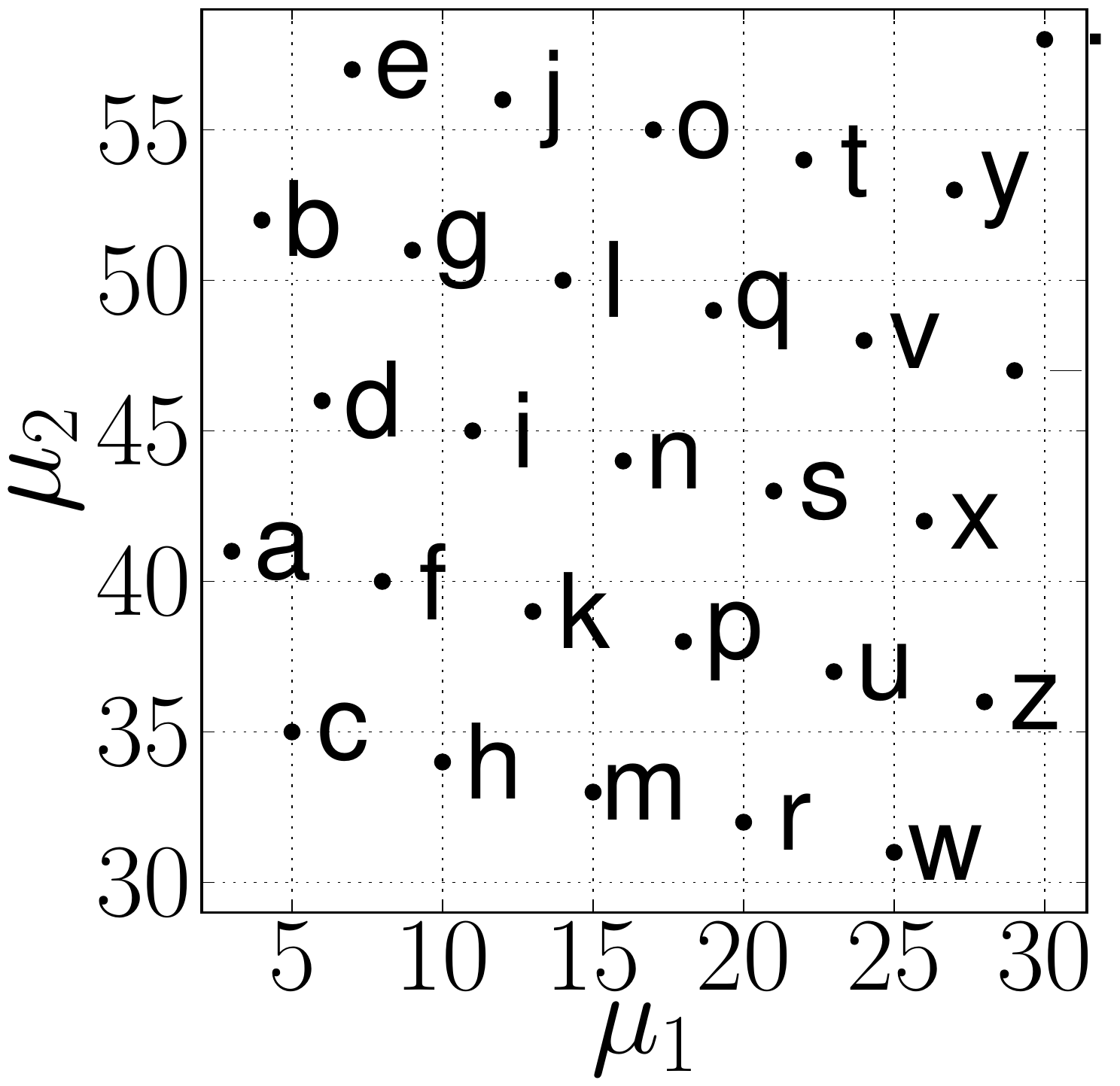} \end{minipage} &
    \begin{minipage}{2.7cm} \includegraphics[width=2.7cm]{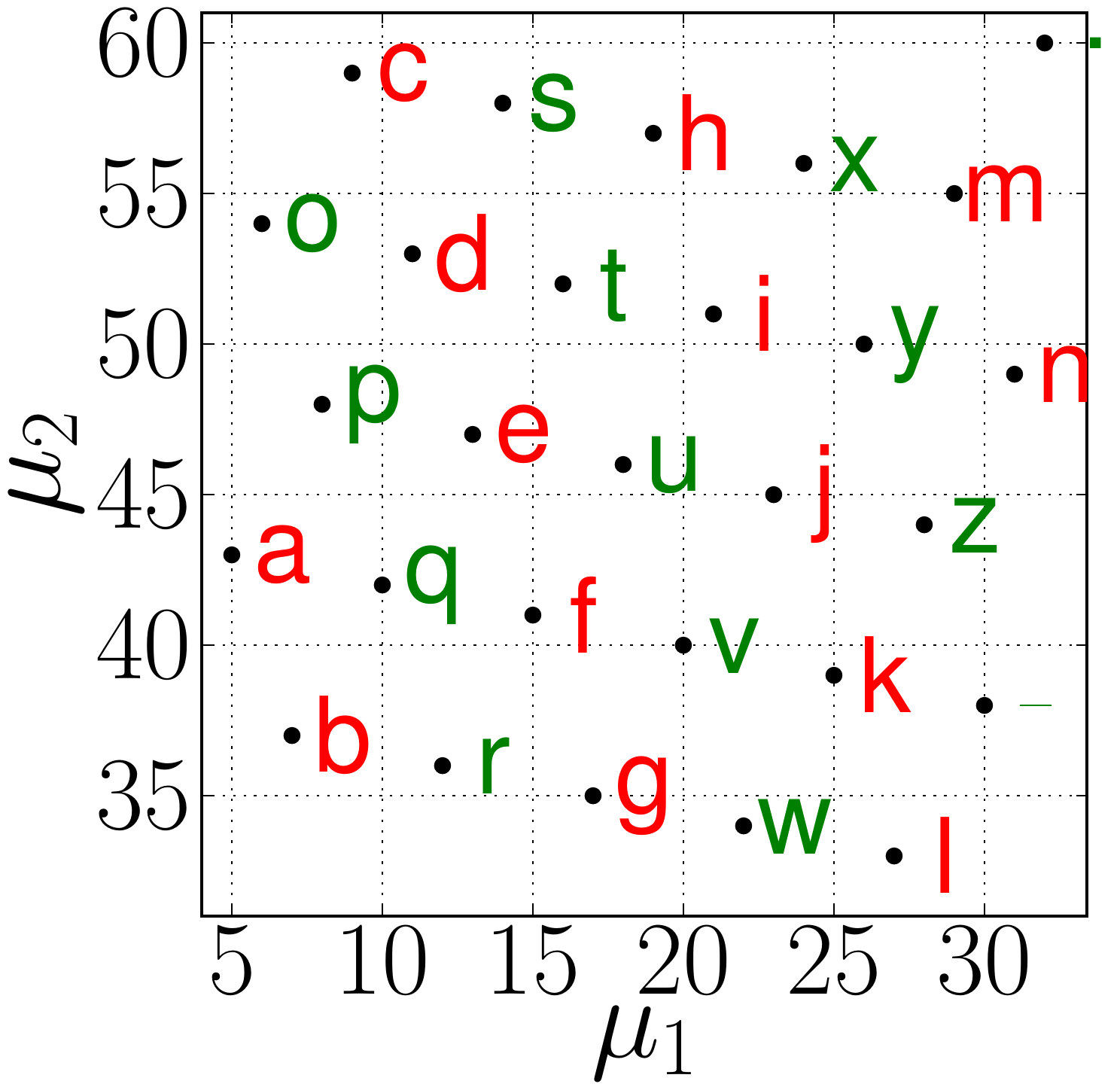} \end{minipage} &
    \begin{minipage}{2.7cm} \includegraphics[width=2.7cm]{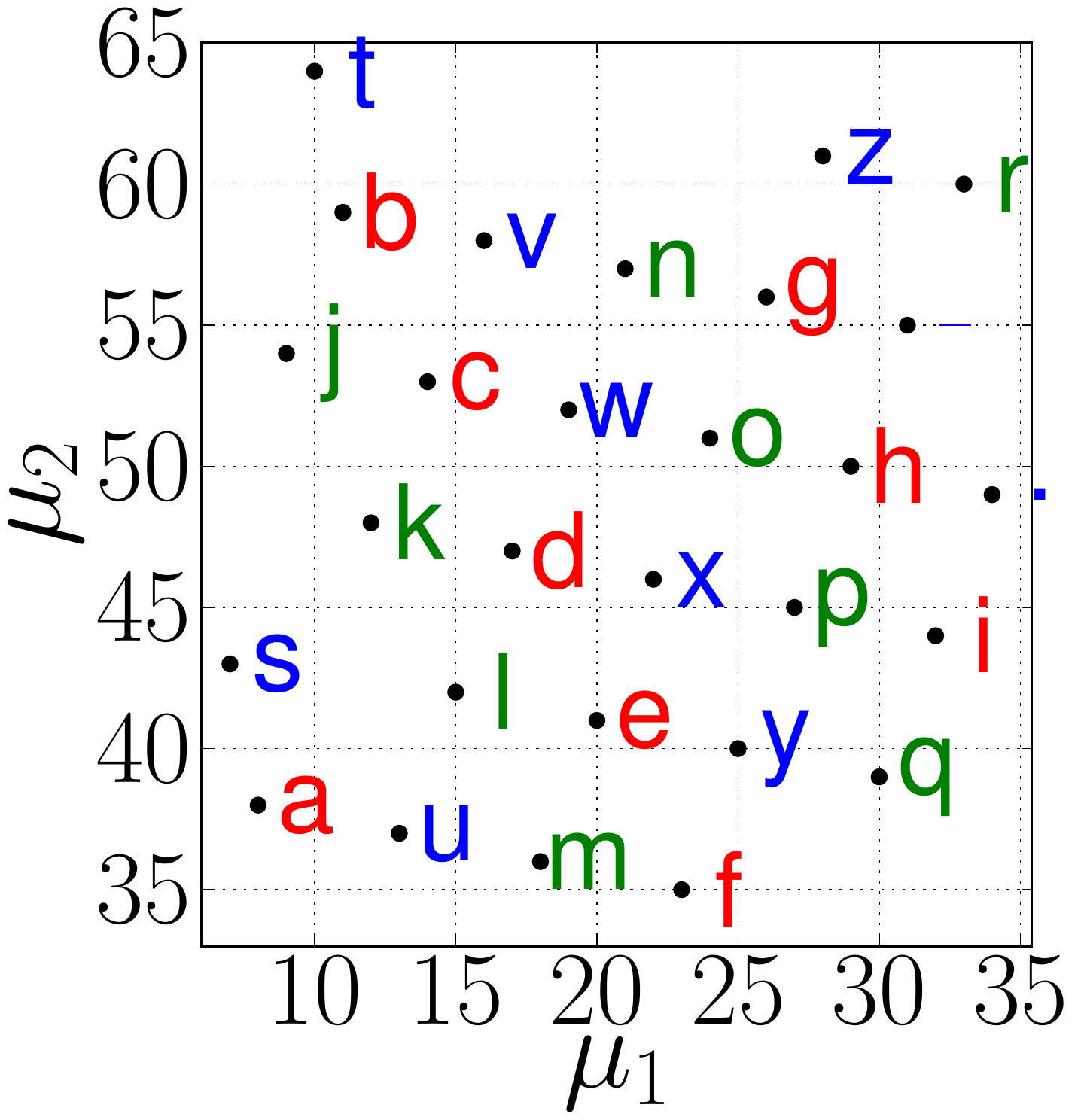} \end{minipage} \\
     {\small 1)} & {\small 2)}  &  {\small 3)} 
    \end{tabular}
    
    \\
    
     \begin{tabular}{c@{}c@{}} 
    \begin{minipage}{2.7cm} \includegraphics[width=2.7cm]{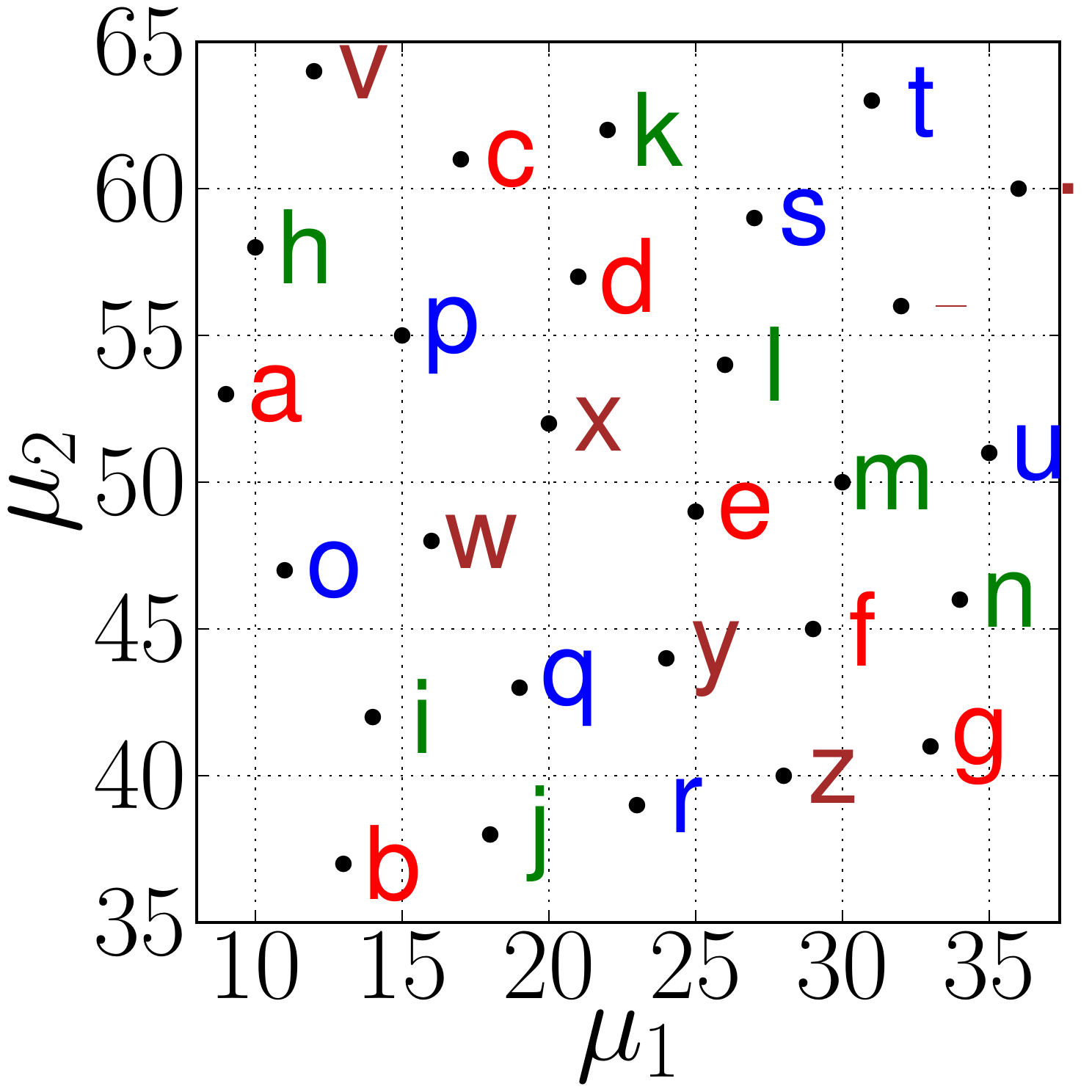} \end{minipage} &
    \begin{minipage}{2.7cm} \includegraphics[width=2.7cm]{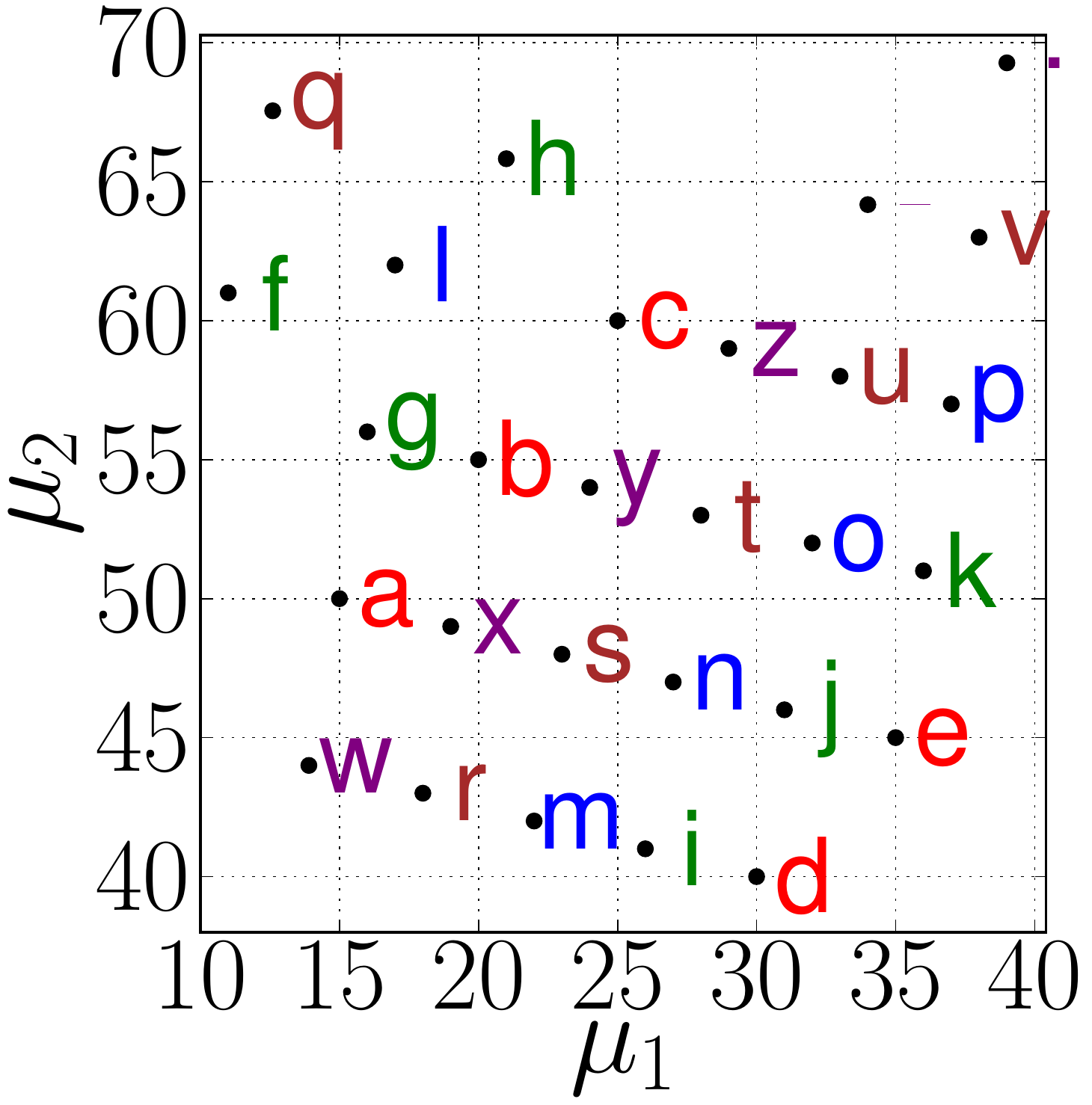} \end{minipage} \\
      {\small 4)}  &  {\small 5)} 
    \end{tabular} 
   
    \\ (a)  \\  
    \begin{minipage}{8.2cm} \includegraphics[width=8.2cm]{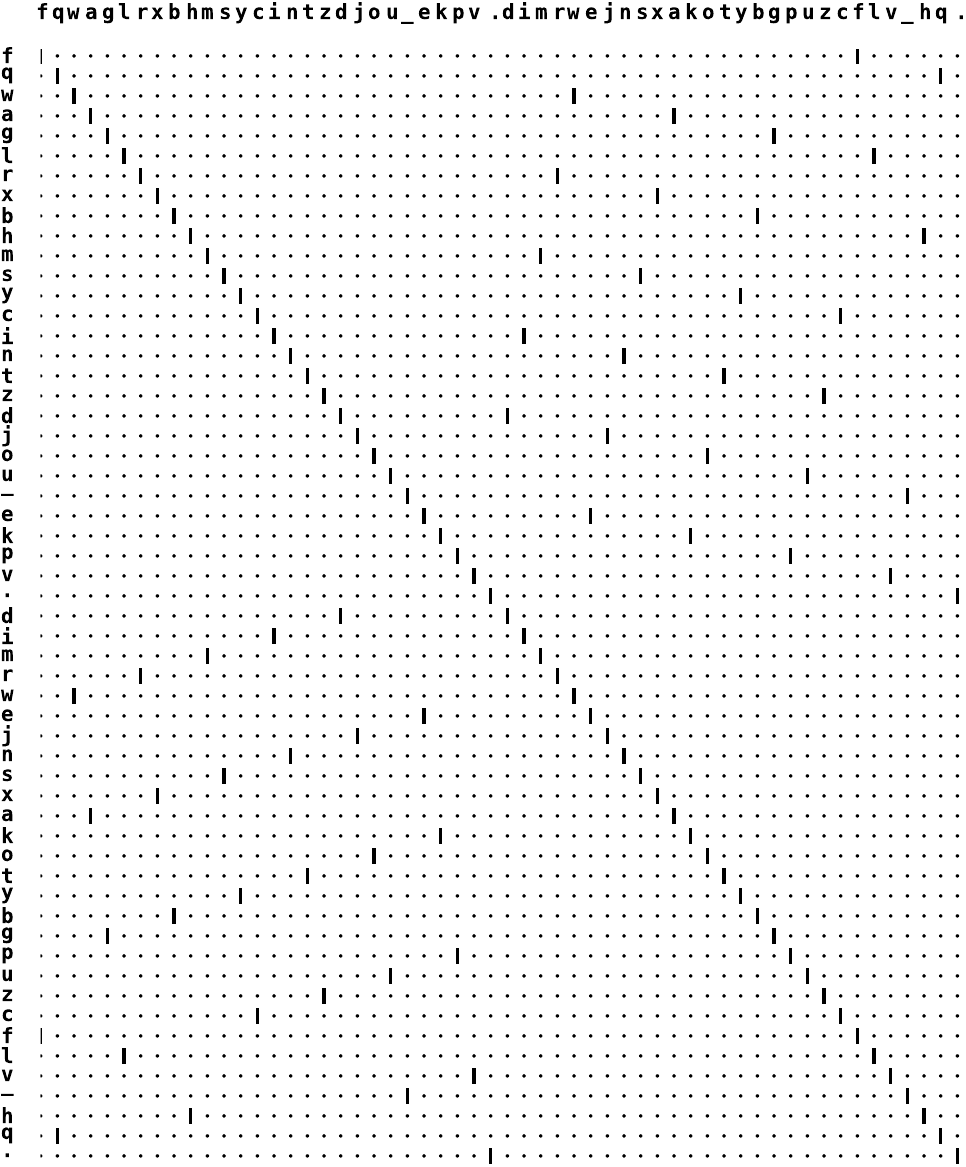} \end{minipage} 
    \\ (b) \\ \hline 
    
   \end{tabular} 
  \caption{(a) The final composite audio sequences for 1-5 channels and their corresponding 2D plots. Letters  from the same clip are represented by the same colour.
  (b) The codewords associated with all letters in the alphabet for  5 channels (as part of computing the composite audio sequence). The y-axis indicates the desired letter, whereas the x-axis indicates which letter to select (``1'') or not to select (``0''). 
 }
  \label{fig:letter_config}
   \end{figure}

   Figure~\ref{fig:letter_config}(a) depicts the final composite audio sequences in 2D for all channels.
  
The optimisation of the composite audio sequence corresponds to maximisation of the information rate~$B$, measured in bits per second:
\begin{equation} \label{eq:channel_capacity}
B = \frac{\mathcal{I}(\mathbf{x} ; \mathbf{y})}{T_{Y}} 
       = \frac{\mathcal{H}(\mathbf{x}) - \mathcal{H}(\mathbf{x} \mid \mathbf{y})}
         {T_{Y}}, 
\end{equation} 
where $\mathbf{x}$ is the input set (a list of words that the user intends to write), $\mathbf{y}$ is the output set (the list of words the user writes), $\mathcal{I}(\cdot)$ 
is the mutual information, 
$\mathcal{H}(\cdot)$ refers to the entropy function, and $T_{Y}$ is the time it takes to produce an output; see~\cite{chi2016} for further detail.

We made some simplifying assumptions to construct our approach.  These assumptions were made to reduce computational cost, and increase generality of use.  
Firstly, we have assumed a Uniform prior for each word, thereby ignoring our dictionary. When one does not ignore the dictionary, it can naturally lead to sequences where
the separation between letters that are frequently close to each other are increased. It may then become much easier to select frequently occurring words, thereby increasing the overall text-entry speed. Note from Figure~\ref{fig:letter_config}(a) that ``o'' and ``e'' are close to each other when considering all sound pairs  in the 5-channel configuration. In English, these two vowels frequently occur next to each other. By default, pairs like ``o'' and ``e'' would therefore limit the performance of the system  by definition, if the user writes in English.

One can also think of Ticker as entering a binary code to write a word. The code becomes longer as the user selects more letters. The intentional word 
can typically be decoded more easily when using longer pseudo random codes, at the expense of a reduction in speed. This idea relates Ticker to Shannon's noisy coding theorem.
Ignoring a dictionary when optimising \eqref{eq:channel_capacity}
reduces the computational complexity considerably, as it reduces the problem to considering only letter codewords (such as shown in Figure~\ref{fig:letter_config}(b)).

Some simplifying assumptions of less significance during the optimisation of the composite audio sequence were: 
 the audio files of all letters were assumed to have the same length, making the denominator in \eqref{eq:channel_capacity} irrelevant.
False positive and false negative switch noise were ignored (assuming
 the user clicks exactly $R$ times for each letter). A rudimentary click-timing model was also assumed.
 
The above simplifications were not applied during inference while using Ticker, but only during the optimisation of the composite audio sequence.
During inference a much more comprehensive noise model is used  to 
allow for more interesting and smooth click-timing models, and which can also adapt to the user.
The letter priors are also not ignored during inference.

\begin{figure*}[!!!!!tbp]
\centering
\begin{tabular}{|c|} \hline
\\
 \begin{minipage}{16cm}  \includegraphics[width=15cm]{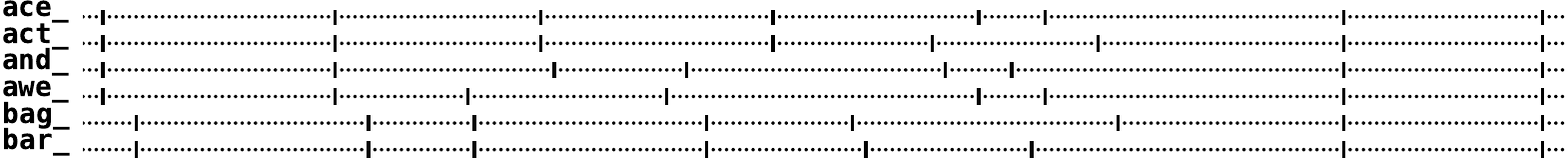}  \end{minipage}  \\ \\  \hline
\end{tabular}           
 \caption{
  A~visualisation of the binary codes associated with some words.}
  \label{fig:word_codes}
\end{figure*}

\figref{fig:word_codes} illustrates that each word in the dictionary
has a unique pseudo-random binary code.
Each letter is associated with a dot or line: The dot represents a zero,
and is associated with the letters that the user should not select
while the composite audio sequence is presented.
The lines represent the timings of the desired letters
(indicating when the user has to click). 
The system is optimal if the distances between confusing,
frequently occurring letters are maximised in binary symbol space.
For example, if the shown codes for \usertext{ace\_} and \usertext{act\_}  have similar (large) prior probability mass,
the ones associated with \qtxt{e} and \qtxt{t} should ideally be
as far from each other as possible. The binary codes for letters (shown in \figref{fig:letter_config}(b))
will then determine if the latter two letters  can be easily distinguished from each other.

\section{The click-timing distribution}

The derivation of the click-timing distribution is shown in \figref{table:tpoe}:
An expression for $P(\mathbf{t}, M, \mathbf{z},   \boldsymbol{\theta}, \ell,  \boldsymbol{\alpha})$ is derived, 
 where  $\mathbf{z}=\{\mathbf{n}, \mathbf{c}, C, N\}$,
 and $\boldsymbol{\alpha}$ is the set of  fixed (untrainable) hyperparameters that controls the distributions over the (trainable) parameters in $\boldsymbol{\theta}$.
 \figref{table:tpoe}(d)--(e) explains
 how to compute the products of Gaussians $p_{\mathbf{z} \mathbf{t} \ell}$. Let
\begin{equation} \label{eq:cum_sum} 
p_{C\ell} 
 = \sum_{\mathbf{c}, \mathbf{n}} p_{\mathbf{z} \mathbf{t} \ell}.
\end{equation} 
Instead of explicitly evaluating all the realisations of $(\mathbf{c},\mathbf{n})$  
(which can quickly become infeasible),  the latter sum can be calculated recursively, 
as summarised in Table~\ref{table:messagepassing}.
Combining \eqref{eq:tpoe} and \eqref{eq:cum_sum}, it then follows that  
\begin{equation} \label{eq:click_prob}
P( \mathbf{t}, M \mid  \boldsymbol{\theta}, {\ell}) = e^{-\lambda T} \sum_{C=0}^{C'} 
 \lambda^{N} 
 \cdot f^{R-C} \cdot (1 -  f)^{C}
 \cdot p_{C\ell} ,
\end{equation}            
where $C'=\min(R,M)$,
      $N=M-C$. 

\begin{figure*}[!!!!htb] 
{\small \setlength{\extrarowheight}{1mm}
\def\arraystretch{1.1}
\begin{tabular}{|m{0.05cm}m{3.7cm}|m{3.5cm}|m{8.5cm}|} \hline 

\multicolumn{4}{|l|}{
\begin{minipage}{16cm}
\begin{equation} \label{eq:tpoe}
P(\mathbf{t}, M, \mathbf{z},   \boldsymbol{\theta}, \ell,  \boldsymbol{\alpha}) = 
\begin{cases}  
 \mathbf{q}  &  \text{if } t_{1} < \ldots < t_{M}, M=N+C  \\
 0, & \text{otherwise}, 
\end{cases} 
\end{equation}
where 
\begin{eqnarray}
 \mathbf{q}  &=& P(\mathbf{t} \mid   \mathbf{z}, \boldsymbol{\theta}, \ell )
\cdot P(\mathbf{c} \mid   C , \boldsymbol{\theta} )
\cdot P(\mathbf{n} \mid N, M  )
\cdot P(M \mid N, C )
\cdot P(N \mid \boldsymbol{\theta}  )
\cdot P(C \mid  \boldsymbol{\theta} )
\cdot P(\ell \mid  \boldsymbol{\theta}  ) \nonumber
\cdot P(\boldsymbol{\theta} \mid \boldsymbol{\alpha}  )
\cdot P(\boldsymbol{\alpha} )
  \label{eq:q}\\
  &=&   p_{\mathbf{z} \mathbf{t} \ell}  
  \cdot e^{-\lambda T} \lambda^{N} 
  \cdot f^{R-C}  (1 -  f)^{C}
  \cdot \pi_{\ell} 
  \cdot \pi_{ \boldsymbol{\theta}} 
  \cdot \pi_{\boldsymbol{\alpha} } \nonumber \\
  &=&   p_{\mathbf{z} \mathbf{t} \ell}  
  \cdot e^{-\lambda T} \lambda^{N} 
  \cdot f^{R-C}  (1 -  f)^{C}
  \cdot \pi_{\ell} 
  \cdot \left [ P( \Delta \mid \beta, \kappa_{\Delta}, \Delta_{0})
\cdot P( \beta \mid a_{\beta}, b_{\beta},   )
\cdot P( \lambda \mid a_{\lambda}, b_{\lambda},   )
\cdot P( f \mid a_{f}, b_{f} ) \right ]
  \cdot \pi_{\boldsymbol{\alpha} },
 \nonumber \\   \label{eq:prior}
\end{eqnarray}
$\mathbf{z}=\{\mathbf{n}, \mathbf{c}, C, N\}$, and 
$ p_{\mathbf{z}\mathbf{t}\ell}$ is defined by Equation~\ref{eq:timeprob}.
\end{minipage}
}

\\  \hline \multicolumn{4}{|c|}{(a)} \\ \hline
      
          &  $  \mathbf{q}=  \displaystyle P(\mathbf{t} \mid   \mathbf{z}, \boldsymbol{\theta}, \ell )$ 
           & \centering $\displaystyle 
             \frac{M!}{T^N C! } \cdot p_{\mathbf{z} \mathbf{t} \ell}  
             $
           & $\mathbf{t}$: Observed click times,
               $t_{m} \in [0,T]$,  $t_{1} < \ldots < t_{M}$,
           \\ 
\cline{4-4}
$\times$   & $\displaystyle P(\mathbf{c} \mid   C , \boldsymbol{\theta} )$
           & \centering $\displaystyle\frac{(R-C)! \, C!}{R!}$
           & $\mathbf{c}$: False negative labels, $c_{r} \in \{0,1\}$, $\sum_{r=1}^{R} c_{r} = C$.
           \\
\cline{4-4}   
$\times$  & $\displaystyle P(\mathbf{n} \mid N, M  )$
           & \centering $\displaystyle\frac{(M-N)! \, N!}{M!}$
           &  $\mathbf{n}$: False positive labels,
              $n_{m} \in \{0,1\}$, $\sum_{m=1}^{M} n_{m} = N$. 
           \\
\cline{4-4}
$\times$   & $P(M \mid N, C )$ 
           & \centering $ \displaystyle \delta( M - (N+C) ) $
           & $R$: The number of times each letter is repeated.
           \\
           
\cline{4-4}
$\times$   & $\displaystyle P(N \mid \boldsymbol{\theta}  )$  
           & \centering $ \displaystyle  \frac{e^{-\lambda T} (\lambda T)^{N}}{N!}$
           & $N$: The number of false positives, $N \in \{0, M\}$.
           \\

\cline{4-4}   
$\times$   & $\displaystyle P(C \mid  \boldsymbol{\theta} )$  
           & \centering $\displaystyle  \frac{f^{R-C} (1 -  f)^{C} R!}{C! \, (R-C)!}$ 
           &  $C$: The number of true clicks,  $C \in \{0, \min(R, M) \}$.  $
            \newline M$: The number of observations, i.e., $M=|\mathbf{t}|=N+C$. 
           \\
\cline{4-4}
$\times$   & $\displaystyle P(\ell \mid  \boldsymbol{\theta}  )$  
           & \centering $\displaystyle \pi_{\ell}$
           & $\ell$: One of $A$ letters in the alphabet.
           \\ \cline{4-4} 
$\times$   & $\displaystyle P(\boldsymbol{\theta} \mid \boldsymbol{\alpha}  )$  
           & \centering $\displaystyle \pi_{\boldsymbol{\theta}}$
           & $\boldsymbol{\theta}$: The collection of  variable (trainable)  parameters.
           \\ \cline{4-4} 
$\times$   & $\displaystyle P( \boldsymbol{\alpha}  )$  
           & \centering $\displaystyle \pi_{\boldsymbol{\alpha}}$
           & $\boldsymbol{\alpha}$: The collection of  fixed (hyper) parameters.
\\ \hline \multicolumn{4}{|c|}{(b)} \\ \hline

           &  $ P(\boldsymbol{\theta} \mid \boldsymbol{\alpha}  )$
           &  \centering $\displaystyle \pi_{\boldsymbol{\theta}}$
           &  The prior over variable (trainable) parameters. \\ \cline{4-4}
           
 $=$       & $\displaystyle P( \Delta \mid \beta, \kappa_{\Delta}, \Delta_{0})$
           & \centering $\displaystyle \mathcal{N}( \Delta \mid \Delta_{0},  (\kappa  \beta )^{-1})$
           & $\Delta$: The average click-time delay (a Gaussian prior). 
           \\ \cline{4-4}   
           
$\times$   & $\displaystyle P( \beta \mid a_{\beta}, b_{\beta},   )$
           & \centering $\displaystyle  \mathrm{Gamma}( \beta \mid a_{\beta}, b_{\beta})$
           & $\beta$: The precision of the click-time delay (Gamma prior), where $\beta=\frac{1}{\sigma^2}$. 
           \\ \cline{4-4}  
$\times$   & $\displaystyle P( \lambda \mid a_{\lambda}, b_{\lambda},   )$
           & \centering $\displaystyle  \mathrm{Gamma}( \lambda \mid a_{\lambda}, b_{\lambda})$
           & $\lambda$: False-positive rate parameter (Gamma prior). 
           \\ \cline{4-4}  
$\times$   & $\displaystyle P( f \mid a_{f}, b_{f} )$
           & \centering $\displaystyle  \mathrm{Beta}(f \mid a_{f}, b_{f})$
           & $f$: False-negative parameter (Beta prior). 
\\ \hline \multicolumn{4}{|c|}{(c)} \\ \hline 

\multicolumn{3}{|l|}
{
\begin{tabular}{l}
\begin{minipage}{7.6cm}
\begin{equation}\label{eq:timeprob}  
  p_{\mathbf{z}\mathbf{t}\ell} 
 =  \prod_{m=1}^{M} \prod_{r=1}^{R}  \mathcal{N}(t_{m}  \mid \boldsymbol{\theta}_{\ell r} )^{g_{\mathbf{z}mr} };
 \end{equation}
 \begin{equation}\label{eq:clicktimeconstraint} 
 g_{\mathbf{z}mr} =  \delta(1-n_{m}) \cdot \delta(1-c_{r}) \cdot \delta( \sum_{m'=1}^{m} \mathbf{n}_{m'} - \sum_{r'=1}^{r} \mathbf{c}_{r'}]) ;
  \end{equation} 
 \end{minipage}
 \\  where $t_{1} < \ldots < t_{M}$ and $\mu_{\ell_{1}} < \ldots < \mu_{\ell_{M}}$. \\
 
\end{tabular}
 }   
  
 & 
 
 {\normalsize 
 \begin{tabular}{c}
\\  
\begin{tabular}{|c@{}c@{}c@{}c|c|c@{}c@{}c@{}c|c|} \cline{1-4} \cline{6-10} 
\multicolumn{4}{|c|}{$\mathbf{n}$}    &                 & \multicolumn{4}{c|}{$\mathbf{c}$}                    &   \multirow{2}{*}{$p_{\mathbf{z}\mathbf{t}\ell}$}            \\ \cline{1-4}\cline{6-9} 
$t_{4}$ & $t_{3}$ & $t_{2}$ & $t_{1}$ &                 & $\ell_{4}$  & $\ell_{3}$ & $\ell_{2}$ & $\ell_{1}$   &                                                                       \\ \cline{1-4}\cline{6-10} 
0       &     0   &  1      & 1       &                 &  0          &     0      &  1         & 1            &   $\mathcal{N}(t_{1} \mid \boldsymbol{\theta}_{\ell_1}) \mathcal{N}(t_{4} \mid \boldsymbol{\theta}_{\ell_2})$ \\   
0       &     1   &  0      & 1       &                 &  0          &     1      &  0         & 1            &   $\mathcal{N}(t_{1} \mid \boldsymbol{\theta}_{\ell_1}) \mathcal{N}(t_{4} \mid \boldsymbol{\theta}_{\ell_3})$ \\   
1       &     0   &  0      & 1       &                 &  1          &     0      &  0         & 1            &   $\mathcal{N}(t_{1} \mid \boldsymbol{\theta}_{\ell_1}) \mathcal{N}(t_{4} \mid \boldsymbol{\theta}_{\ell_4})$ \\  \cline{1-4}
0       &     1   &  1      & 0       &$\longrightarrow$&  0          &     1      &  1         & 0            &   $\mathcal{N}(t_{1} \mid \boldsymbol{\theta}_{\ell_2}) \mathcal{N}(t_{4} \mid \boldsymbol{\theta}_{\ell_3})$ \\  \cline{1-4}
1       &     0   &  1      & 0       &                 &  1          &     0      &  1         & 0            &   $\mathcal{N}(t_{1} \mid \boldsymbol{\theta}_{\ell_2}) \mathcal{N}(t_{4} \mid \boldsymbol{\theta}_{\ell_4})$ \\  
1       &     1   &  0      & 0       &                 &  1          &     1      &  0         & 0            &   $\mathcal{N}(t_{1} \mid \boldsymbol{\theta}_{\ell_3}) \mathcal{N}(t_{4} \mid \boldsymbol{\theta}_{\ell_4})$ \\  \cline{1-4} \cline{6-10}
\end{tabular}
\\
\end{tabular}
}

 \\ \hline \multicolumn{3}{|c|}{(d)} & \multicolumn{1}{ c|}{(e)}\\ \hline

\end{tabular}
}
\caption{(a) To do inference in Ticker \eqref{eq:tpoe} has to be computed. It is factorised in \eqref{eq:q}. The first column of (b)  contains the same factorisation, with
the values of the probabilities shown in the second column, and 
 a description of each term in
the third column. For example,  $  P(\mathbf{t} \mid   \mathbf{z}, \boldsymbol{\theta}, \ell ) =  
             \frac{M!}{T^N C! } \cdot p_{\mathbf{z} \mathbf{t} \ell}  $. 
(c) The factorisation of the prior $P(\boldsymbol{\theta} \mid \boldsymbol{\alpha})$ is displayed in a similar way. 
\eqref{eq:q} is derived by multiplying all the terms in the second column of (b). 
 Likewise $\pi_{ \boldsymbol{\theta}}$  in \eqref{eq:prior} is derived by multiplying all the terms in the second column of (c).
A justification of all the models in the second column of (b) and (c) is provided in the text.
 (d) $ p_{\mathbf{z}\mathbf{t}\ell}$ is defined by Equation~\ref{eq:timeprob}.
 (e) Example realisations of $p_{\mathbf{z} \mathbf{t}\ell}$ defined in (d) with $M=4$, $N=2$, $C=2$, $R=4$.
Each realisation of $\mathbf{n}$ corresponds to labelling each of the received click times as either 
a false/true positive. For example, the highlighted $\{0,1,1,0\}$ indicates that 
$t_{1}$ and $t_{4}$ are true clicks. For each such realisation of $\mathbf{n}$, all possible
realisations  of $\mathbf{c}$ are considered which determines the
product of Gaussians that should be used (all products where $t_{1}$ and $t_{4}$ are true clicks). }
\label{table:tpoe}
\end{figure*}

 \begin{table*}[!!!!htb]
 \centering
\begin{tabular}{|l|} \hline  
{\bf Initialise}:
\\
\begin{minipage}{15cm}
\begin{itemize}
\item For each $\ell$, construct a matrix $G_{\ell}$ of size $M \times R$, where 
$g_{\ell}(m,r) = \mathcal{N}(t_{m} \mid \boldsymbol{\theta}_{\ell_{r}})$. 
\item Construct a matrix $\boldsymbol{\alpha}_{C}$ of size $(M+1) \times (R+1)$, where 
$\alpha_{C}(m,r)=1$, $\forall (m,r)$. 
\end{itemize}
\end{minipage}
\\

{\bf For $C = \{1, \ldots, \min(R,M)\}$}:
\\

\begin{minipage}{15cm}
\begin{itemize}
\item[] {\bf For $r = \{R', \ldots, 1\}$}:
\begin{itemize}
\item[]  {\bf For $m = \{M', \ldots, 1\}$}:
\begin{itemize}
\item[] $\alpha_{C}(m,r) = g_{\ell}(m,r) \cdot \alpha_{C-1}(m+1,r+1) + \alpha_{C}(m,r+1) + \alpha_{C}(m+1,r)  $    
\end{itemize}
\end{itemize}
\item[] $p_{C\ell} = \alpha_{C}(1,1)$
\end{itemize}
\end{minipage}
\\
\begin{minipage}{15cm}
where 
\begin{itemize}
\item $M'=M-C+1$ and $R'=R-C+1$. 
\item At the boundaries, $\alpha_{C}(M'+1,r)=0$ and $\alpha_{C}(m,R'+1)=0$.  
\end{itemize}
\end{minipage}
\\ \hline
\end{tabular}
\caption{The recursion algorithm for \eqref{eq:cum_sum}. 
 Note that the size of $\boldsymbol{\alpha}_{C}$ decreases to  $M' \times R'$ as $C$ increases.
 }
\label{table:messagepassing}
\end{table*}

\section{ Training the Click-Timing Model} 
 \label{subsec:adapt}

Online adaptation of the model is done based on the last $w_{\mathrm{offset}}$ word selections corresponding to the letter sequence $\{ \ell^{*}_{1}, \ldots, \ell^{*}_{H}\}$.
This letter sequence corresponds to a sequence of received click-time ensembles
$\{ \{ \mathbf{t}^{*}_{1}, M^{*}_{1}\}, \ldots, \{ \mathbf{t}^{*}_{H}, M^{*}_{H}\}\}$.
An online-learning rate is included to limit the influence of erroneous word selections. 

 The E-M algorithm is specifically used to compute the maximum-a-posteriori (MAP) estimates
$\boldsymbol{\theta}^{*}$ of our parameters $\boldsymbol{\theta}$.
 Only true positive click times should contribute to the kernel-density estimation, which results in the following mixture model:
\begin{equation}\label{eq:kernel_density_generic}
 P(t_{m} \mid \mathbf{z}, \ell, \mathbf{t}^{*}, \sigma_{\mathrm{K}} )
   =  \frac{ \displaystyle \sum_{h,\mathbf{z}^{*}}
        \frac{\pi_{h\mathbf{z}^{*}}}{\sigma_{\mathrm{K}}}
        \mathcal{N}\left(\frac{ t_{m\ell\mathbf{z}}-t^{*}_{h\mathbf{z}^{*}}}{\sigma_{\mathrm{K}}} \mid 0, 1 \right )
      }{ \displaystyle \sum_{h,\mathbf{z}^{*}} \pi_{h\mathbf{z}^{*}} }, 
\end{equation}
 where $t^{*}_{h\mathbf{z}^{*}}$ corresponds to the normalised click-time associated with 
$\ell_{h}^{*}$ (already selected).  Each true positive is weighted by $\pi_{h\mathbf{z}^{*}}$, 
where, by definition, $\pi_{h\mathbf{z}^{*}} \in (0, 1)$. 
 Each hypothesis contained in $\mathbf{z}^{*}$ stipulates which repetition  
 $r^{*}$ of $\ell_{h}^{*}$ is responsible for a positive. 
 This
enables click-time normalisation, which involves subtracting the corresponding starting time of the sound file from the stored letter time.  
  Likewise, the newly received click-time $t_{m}$ is 
normalised by subtracting the starting time of audio file
corresponding to $\ell$ and $r$ that is specified by $\mathbf{z}$.
  
It is computationally expensive to train  
$\sigma_{\mathrm{K}}$.
A well known approximation amounts to firstly approximating the non-parametric distribution with a Gaussian distribution (by using the E-M algorithm in our case). 
Secondly, the standard deviation of the latter Gaussian is scaled 
according to the {\em normal-scale rule}, in order to compute $\sigma_{\mathrm{K}}$~\cite{silverman1986density}.

  The training procedure is followed every-time a new word is selected. Table~\ref{table:mstep} summarises the E-M update equations for our application, where the same generic notation defined in~\cite{Bishop2006} were used to do the derivations.  At convergence, the Gaussian click-time  parameters $\{\Delta, \sigma\}$ and the switch noise parameters $\{f, \lambda\}$ are  set. These point estimates~$\boldsymbol{\theta}^{*}$ of the parameters are regularised by the fixed hyperparameters~$\boldsymbol{\alpha}$ provided by Figure~4.

 \begin{table}[!!!!htbp] 
{\normalsize \setlength{\extrarowheight}{1mm}
\begin{tabular}{|p{8cm}|} \hline 
{\bf Expectation: } \\ \hline
\begin{minipage}{8cm}
\begin{equation}\label{eq:yz}
 \gamma_{\hat{\mathbf{z}}h}  = P(\hat{\mathbf{z}}  \mid  \mathbf{t}_{h},  \ell_{h},  \boldsymbol{\theta}^{\mathrm{old}}, \boldsymbol{\alpha} ),  
\end{equation}
which can be derived from  Equation~3 for for all the observed letters $\{\ell_{1}, \ldots, \ell_{H} \}$, with $\pi_{\ell_{h}}=1$. The expected number of true clicks is given by: 
\end{minipage}
\\
\begin{minipage}{8cm}
\begin{equation}\label{eq:y_sum} 
 c^{*} = \sum_{h=1}^{H} \sum_{\hat{\mathbf{z}}}   \gamma_{\hat{\mathbf{z}} h},
\end{equation} 
where   $c^{*} \in[0, RH]$.
\end{minipage}
\\\\\hline 

{\bf Maximisation: } \\ \hline

\begin{minipage}{8cm}
\begin{equation}\label{eq:em_newparams} \nonumber
 \boldsymbol{\theta}^{\mathrm{new}} = \underset{ \boldsymbol{\theta}}{\text{arg max }} \mathcal{Q}(\boldsymbol{\theta}, \boldsymbol{\theta}^{old}), \text{ where}
\end{equation} 
\begin{equation}\label{eq:em_cost} \nonumber
 \mathcal{Q}(\boldsymbol{\theta}, \boldsymbol{\theta}^{old}) = \ln \pi_{\boldsymbol{\theta}} + \sum_{h=1}^{H} \sum_{\hat{\mathbf{z}}} \gamma_{\hat{\mathbf{z}}h}
 \ln P(\mathbf{t}_{h}, \hat{\mathbf{z}} \mid \boldsymbol{\theta}, \ell_{h},  \boldsymbol{\alpha}), 
\end{equation} 
and $P(\mathbf{t}_{h}, \hat{\mathbf{z}} \mid \boldsymbol{\theta}, \ell_{h},  \boldsymbol{\alpha})$ can be derived from Equation~3. 
Following the maximisation of $\mathcal{Q}(\boldsymbol{\theta}, \boldsymbol{\theta}^{old})$,
 \end{minipage}
\\ 
\begin{minipage}{8cm}
\begin{equation}\label{eq:Delta}  
\Delta = \frac{\displaystyle \kappa\Delta_{0} + \Delta^{*} 
 }{\displaystyle \kappa + c^{*}  },
\end{equation}
\begin{equation}\label{eq:sigma} 
\sigma^{2} = \displaystyle  \frac{ 2b_{\beta} + \Delta^{**} + \kappa \Delta_{0}^{2}
- \Delta^{2}(\kappa + c^{*} ) }{  2a_{\beta} -1 + c^{*}}
\end{equation}  
\begin{equation}\label{eq:fp_rate} 
\lambda = 
\frac{\displaystyle a_\lambda - 1 +  M^{*} - c^{*}}
{ \displaystyle b_{\lambda}  + TH},
\end{equation}
\begin{equation}\label{eq:fr} 
f = 
\frac{\displaystyle RH + a_{f} - 1   - c^{*} }
{ \displaystyle  RH + a_{f} + b_{f} - 2 }, 
\end{equation}
where  
\begin{equation}\label{eq:delta1} 
\Delta^{*} =  \sum_{h=1}^{H} \sum_{\hat{\mathbf{z}}} \gamma_{\hat{\mathbf{z}}h} \cdot \Delta^{*}_{\hat{\mathbf{z}}h}, 
\end{equation}
\begin{equation}\label{eq:delta2} 
\Delta^{**} =  \sum_{h=1}^{H} \sum_{\hat{\mathbf{z}}} \gamma_{\hat{\mathbf{z}} h} \cdot (\Delta^{*}_{\hat{c}\hat{\mathbf{z}}h})^{2}, 
\end{equation}
\begin{equation}\label{eq:M_sum} 
 M^{*}  = 
\sum_{h=1}^{H} M_{h},
\end{equation}
\begin{equation}\label{eq:delta_aux}  
 \Delta^{*}_{\hat{\mathbf{z}}h}  =  t_{\hat{\mathbf{z}}h} - \mu_{\hat{\mathbf{z}}h}, 
\end{equation} 
where  $M_{h}$ is the number of clicks observed during the selection of letter $\ell_{h}$, $t_{\hat{\mathbf{z}}h}$ is the observed click time,
and  $\mu_{\hat{\mathbf{z}}h}$ is beginning of the audio file implied by the labelling $\hat{\mathbf{z}}$.  
\end{minipage}
\\ \\  \hline

{\bf Kernel bandwidth parameter: } \\ \hline
\begin{minipage}{8cm}
\begin{equation}\label{eq:kernel_weights}
\pi_{ h\mathbf{z}^{*} } = \frac{\gamma_{\hat{\mathbf{z}}h}}{c^{*}}
\end{equation}
\begin{equation}\label{eq:scale}  
\sigma_{\mathrm{K}} \approx 
 \frac{1.06\sigma}{(c^{*})^{0.2}} 
\end{equation} 
 \end{minipage}
\\ \hline \end{tabular}
}
\caption{The E- and M-steps of the E-M algorithm, used to train the parameters of Ticker's noise distributions. Following the convergence of $\mathcal{Q}(\boldsymbol{\theta}, \boldsymbol{\theta}^{old})$, Equations~\ref{eq:kernel_weights}-\ref{eq:scale} are applied once as the last step of training.
}
\label{table:mstep}
\end{table}

   The hyperparameters were chosen to allow for a broad range of parameters to be learned. 
     Their effect also naturally wears off as more training  samples accumulate.
    We  specifically used $a_\lambda = 1.5$,
      $b_\lambda = 60$,
      $a_f= 2$,
      $b_f=10$,
      \nu{$\Delta_{0}=0.1$}s,
      $\kappa=0.01$,
      $a_{\beta} =2$, and
      $b_{\beta}=0.001$
    during the application of the E-M algorithm in our simulations and final user trials.

    After applying the E-M algorithm, each of the parameters  listed in the M step are updated according to the rule $\theta= (1-\lambda_\mathrm{learn})\theta_{\mathrm{old}} + \lambda_\mathrm{learn} \theta_{\mathrm{new}}$. After this step, the normal-scale rule is applied to compute $\sigma_{\mathrm{K}}$. $ \gamma_{\hat{\mathbf{z}}h} $ from the E step is then recalculated to compute $\pi_{ h\mathbf{z}^{*} }$ (see the bottom of  Table~\ref{table:mstep}). To prevent the system becoming too slow, we only use the last 1000 selected letters during training and evaluation.
    
In Nomon~\cite{nomon}, there are fewer latent variables, allowing for a straightforward application of the normal-scale rule. Online learning is applied by using linear dampening to reduce the effect of previous samples when updating the kernel-density estimator. This effectively uses an exponential distribution to model the importance of previous samples: As time progresses the importance of older samples will decay exponentially, allowing them to be pruned in a natural way. In our case, older samples are considered just as important as the newest of samples during evaluation, implying a Uniform distribution. Future work will involve testing the effect of dampening of older samples in the same way as Nomon~\cite{nomon}.      
 
Some shortcomings inherent to MAP estimation need to be considered. Since MAP estimates are variable under a change of basis, one 
 may achieve better results in some cases by applying a non-linear transform to the basis of the probability distribution that models the parameter at hand~\cite{MacKay98}. This option has not been explored for this application.

 Like all MAP estimates,  successful training strongly depends on good initialisation,
 which is done as follows.
First, $\lambda$ and $f$ are initialised by measuring the switch noise. The user is then requested to write the word  ``yes\_" during a calibration phase before starting to use the system for the first time. The E-M procedure is then used to train $\{\Delta, \sigma\}$ while keeping $\{f, \lambda\}$ fixed (to the measured values), and setting  $\lambda_\mathrm{learn}=1.0$. After calibration   $\lambda_\mathrm{learn}=0.3$ and $\{f, \lambda\}$ are updated according to the words that are selected. Note that most switch manufacturers specify the typical false positive- and negative rate in their documentation, so that they don't have to be measured. A fairly large degree of flexibility in the measurement accuracy is also automatically allowed through the hyperparameters.

 In an ideal Bayesian world, one would integrate out all the parameters
 (the trainable ones in $\boldsymbol{\theta}$), instead of inferring point estimates of them.
  However, one should
  consider the effect of this approximation for each application, which is more severe if there is a lack of training data. We have only a few parameters, and the distributions over them are quite simple, which implies that, in our case, we have ample training data: On average, each word selection leads to 
  about ten click-times that can be used during training.  The error bars (standard deviations of the parameter estimates) in many cases decrease with the familiar scale factor $1\sqrt{C_\mathrm{total}}$~\cite{MacKay03}, where $C_\mathrm{total}$ is the total number of click-times during training. Thus, 10 initial click times, which quickly accumulates to at least 100, in addition to the relatively slow learning rate of $\lambda_\mathrm{learn}=0.3$, and our calibration step are considered adequate steps  to ensure ample training data.

In practice it was found that our training procedure tends to cause over smoothing, which limits entry rate. This is a well-known drawback of the  
kernel-density estimation described above, and becomes more prevalent when the distribution is multimodal, since $\sigma$  can become quite large, 
causing $\sigma_{\mathrm{K}}$ to become large. 

In Nomon~\cite{nomon}, it is easier to learn how to click precisely since the procedure (click when the rotating hand reaches noon)  is the same for each letter. 
 Firstly, when using Ticker, the user has to get used to the rhythm within a channel. The rhythm created within the channel is imprecise compared to Nomon~\cite{nomon}. The time at which a sound becomes clearly audible might differ from the ideal click-time (determined by the rhythm) with several tens of milliseconds. 
Secondly, it is more difficult to time the letters at the beginning of the composite audio sequence, even with the added ``tick'' sound since the user has to tune in on a specific voice. By construction, it can therefore be expected that the click-timing distributions in Nomon~\cite{nomon} will be more unimodal than in Ticker, making over-smoothing less of a problem. It is, however, better to have some probability mass associated with all modes in the click-timing distribution, than to have only a unimodal distribution.

A future improvement in Ticker may involve choosing a different click-timing distribution.
A Dirichlet process could be a possible choice.

\section{Language Modelling} 
\label{sec:language}

Let there be $D$ words in the dictionary,
and let the $d$th word $\mathbf{w}_{d}$ consist of a sequence of $S$ letters: 
$\mathbf{w}_{d} =\{\ell_{d1}, \ldots, \ell_{dS}\}$. When necessary,  the notation $\ell_{dsr}$ is used
to refer to the $r$th repetition of letter $\ell_{ds}$. The set of word priors is represented by
 $\Pi_{0} = \{ \boldsymbol{\pi}_{1}, \ldots, \boldsymbol{\pi}_{D}\}$, where $\boldsymbol{\pi}_{d}$ is the normalised frequency 
of the $d$th word in the dictionary.

A counting variable~$k$  keeps track of the letter index.
If at least one click is received  at the end of the composite audio sequence, 
the system will move on to the next letter, and $k$ will be incremented. 
For example, if the user wants to write the word \qtxt{is\_}, $k=1$,
and the user will start by selecting \qsym{i} twice during the presentation
of the composite audio sequence.
If the system received clicks at the end of the composite audio sequence,  
the posterior probabilities of all words in the dictionary are updated, and the counter will be incremented so that $k=2$.
The updated probabilities then become the word priors while $k=2$.
This update procedure is formalised through Bayes' rule,
which provides new posterior word probabilities each time $k$ is updated:
\begin{eqnarray}
\boldsymbol{\pi}_{k,d} &=& P(\mathbf{w}_{d} \mid  k, \{\mathbf{t}_1, M_{1}\}, \ldots, \{\mathbf{t}_k, M_{k}\}, \boldsymbol{\alpha})
                           \nonumber\\  
                       &=& \int_{\boldsymbol{\theta}} P(\mathbf{w}_{d} \mid \boldsymbol{\theta}, \mathcal{D}, \boldsymbol{\alpha}) 
			   P(\boldsymbol{\theta} \mid \mathcal{D}, \boldsymbol{\alpha}) d\boldsymbol{\theta}
                           \label{eq:bayes_words_exact}
\end{eqnarray}
where $\mathcal{D}$ denotes the data $\{k, \{\mathbf{t}_1, M_{1}\}, \ldots, \{\mathbf{t}_k, M_{k}\}\}$.
\eqref{eq:bayes_words_exact} can be approximated as follows:
\begin{eqnarray}
  \boldsymbol{\pi}_{k,d}
  &\approx&
     P(\mathbf{w}_{d} \mid \boldsymbol{\theta}^{*}, \mathcal{D}, \boldsymbol{\alpha}) 
     P(\boldsymbol{\theta}^{*} \mid \mathcal{D}, \boldsymbol{\alpha})  \nonumber \\
  &=&
     \frac{P(\mathbf{t}_k, M_{k} \mid  \boldsymbol{\theta}^{*}, \ell_{d s}) \boldsymbol{\pi}_{k-1,d}}
     {\displaystyle\sum_{d'=1}^{D} P(\mathbf{t}_k, M_{k}\mid  \boldsymbol{\theta}^{*}, \ell_{d' s'}) 
     \boldsymbol{\pi}_{k-1,d'}  } \label{eq:bayes_words}, 
\end{eqnarray}
where $\boldsymbol{\theta}^{*}$ is a point estimate of the parameters, and  the letter indices are given by 
\begin{equation}
\begin{split}
    s  & \ =\   k - \left\lfloor \frac{k-1}{|\mathbf{w}_{d}|} \right\rfloor \cdot |\mathbf{w}_{d}| \\
    s' & \ =\   k - \left\lfloor \frac{k-1}{|\mathbf{w}_{d'}|} \right\rfloor \cdot |\mathbf{w}_{d'}|.
\end{split}
\end{equation}
The letter index~$s$ depends on $k$ and $d$, and allows $\boldsymbol{\pi}_{k,d}$ to be updated 
even if $k > |\mathbf{w}_{d}|$.

For example, if the system has to compute the denominator in \eqref{eq:bayes_words}
for the word \qtxt{is\_} and $k=4$, 
then $s=1$, so that $\ell_{d 1}=\sym{i}$.
 $P(\mathbf{t}_4, M_{4} \mid \boldsymbol{\theta}, \ell=\sym{i})$ is used in \eqref{eq:bayes_words}. 
 The point estimate
$\boldsymbol{\theta}^{*}$ is updated after a word has been selected - more detail about this approximation method is given in Section~\ref{subsec:adapt}.

If $\max(\Pi_{k})$ is bigger than a predefined threshold
for updating \eqref{eq:bayes_words},
$\mathbf{w}_{d}^{*}$ is selected, where 
$\mathbf{w}_{d}^{*} =  \underset{\mathbf{w}_{d}}{\argmax}(\Pi_{k})$.
The selection threshold was chosen to be 0.9 during all the user trials and simulations.
That is, if the system is at least 90\% certain of the intentional word, it will be selected.
Note that using the maximum posterior value in this way is just a heuristic,
and there may well be better alternatives.
 Other heuristics such as the sum of the posterior probabilities of the top few words may be used, for example.

\section{A Case study}

A case study with a non-speaking individual with motor disabilities who was unable to communicate on his own using the standard scanning system Grid2~\cite{grid2} was done. 
This user communicated mostly by raising his eyebrows in an interactive conversation with his carer.
The carer could also guess well what he tried to say after he selected a few letters.
We automated this process using an Impulse switch attached to the user's eyebrow muscle
and connected to Ticker.

The Impulse switch is quite prone to false positives and drift, especially if the user communicates for a while and his body temperature 
slightly increases.
Since this end-user had vision problems all visual cues had to be replaced with audio cues. 

We trained the end-user to use Ticker in four 2-hour sessions.
During the last session the end-user was able to select 20 words (four phrases) at a rate of 1.3~wpm. No time-out errors occurred, and four of the 20 words were wrong. However,  due to the context one could easily see which words the end-user meant. For example, ``throb\_''  were selected instead of ``three\_"  from the phrase 
``three\_two\_one\_zero\_blast\_". All the other words were selected correctly.   
 
\begin{figure*}[!!!!!htbp]
  \centering
  \begin{tabular}{cc} 
    \begin{minipage}{7.5cm} \centering \includegraphics[width=7.5cm]{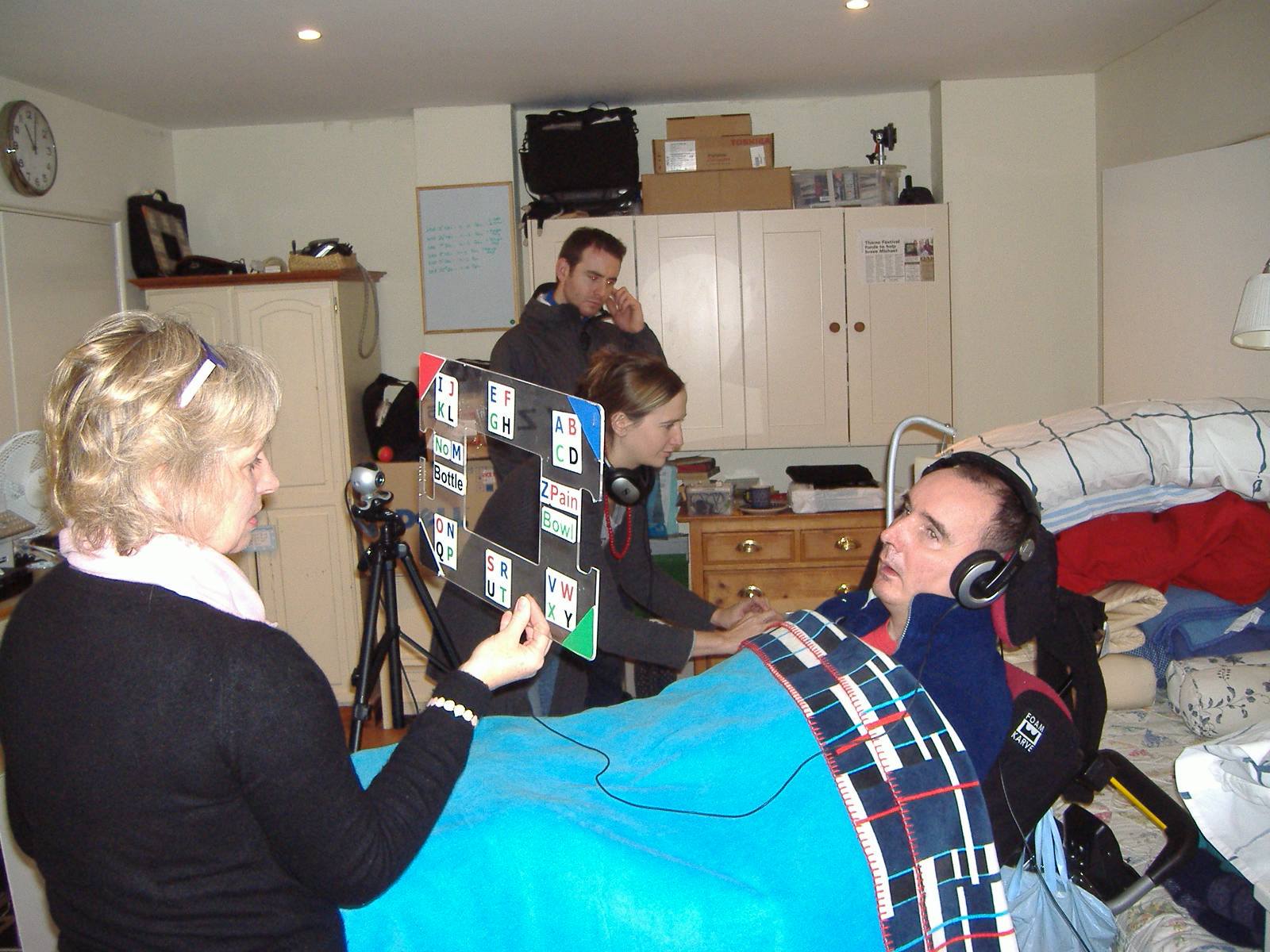}  \end{minipage}   
	&
   \begin{minipage}{7.5cm} \centering \includegraphics[width=7.5cm]{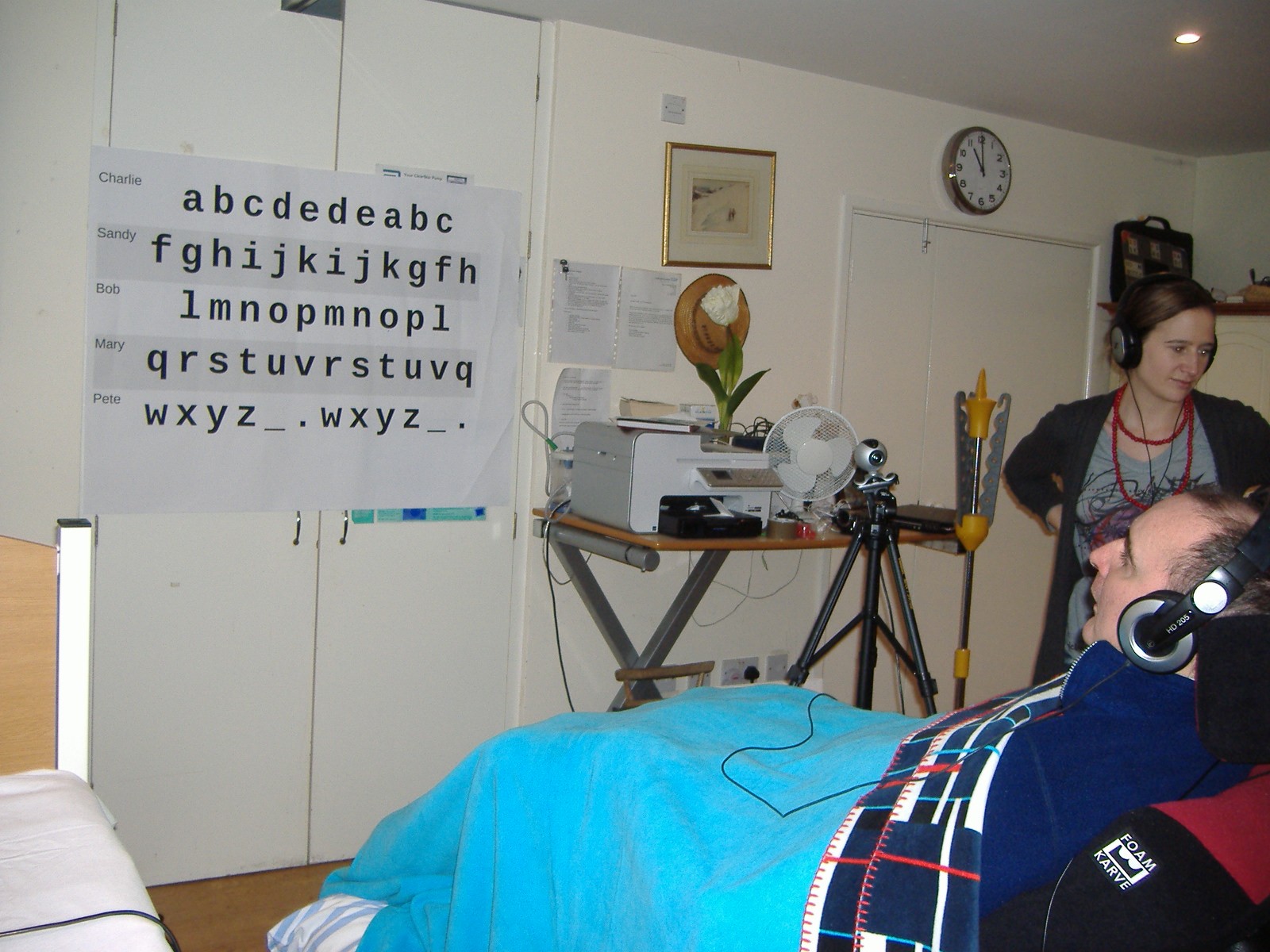}  \end{minipage}   
	\\ (a) & (b) \\
   \begin{minipage}{7.5cm} \centering \includegraphics[width=7.5cm]{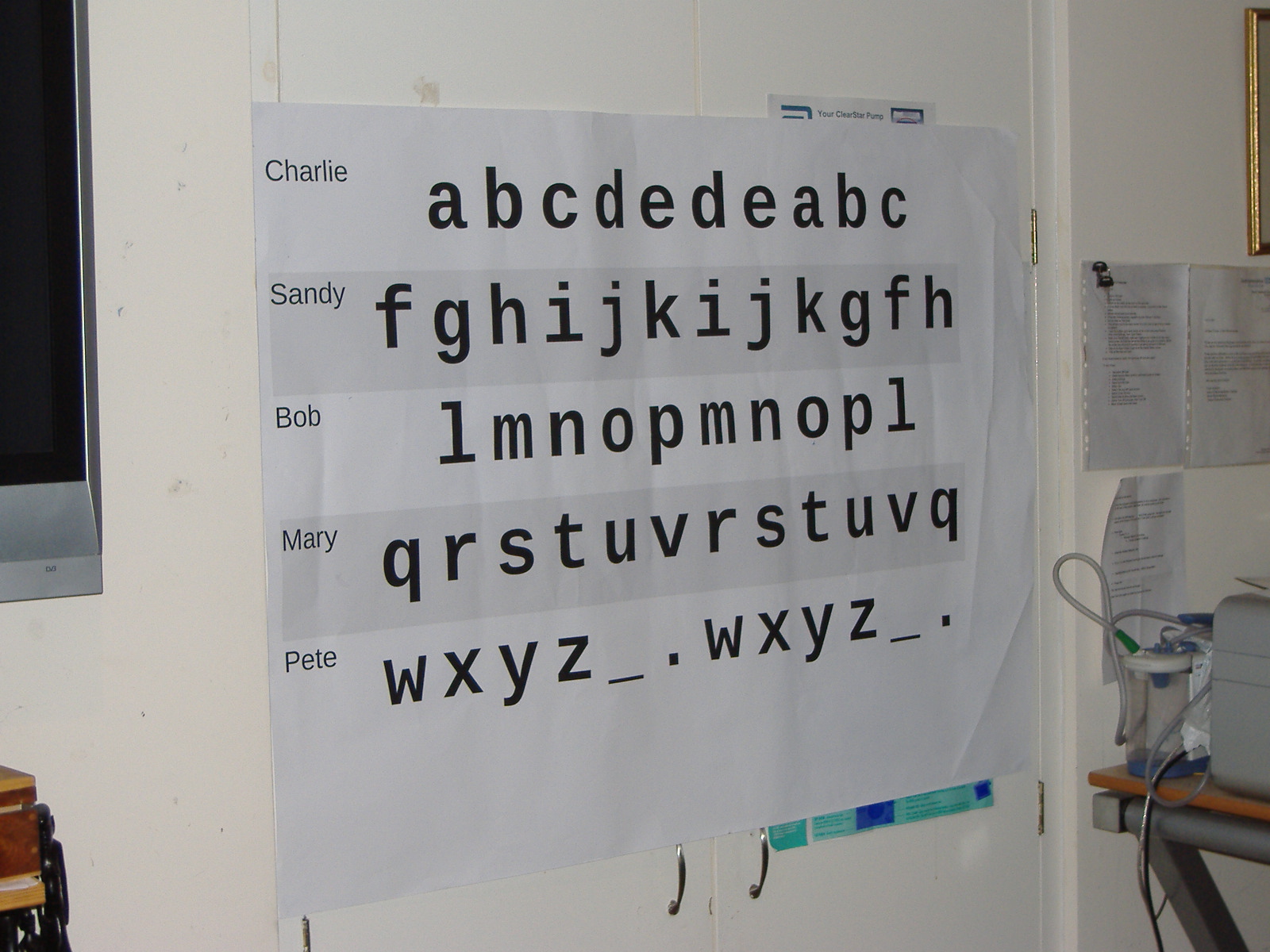}  \end{minipage}  
   	&
   \begin{minipage}{7.5cm} \centering \includegraphics[width=7.5cm]{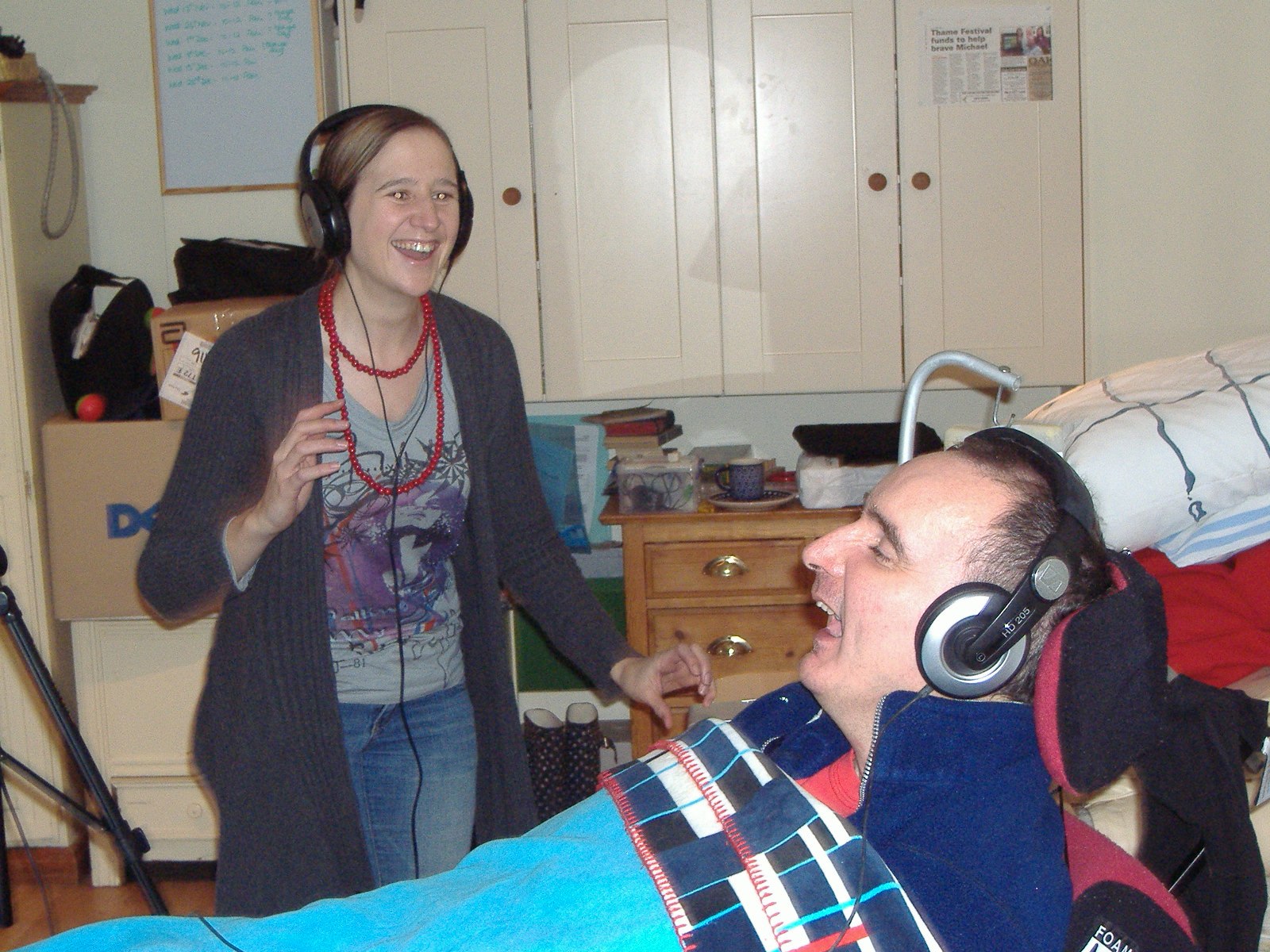}  \end{minipage}    
        \\ (c) & (d) \\
 \end{tabular}

  \caption{%
  (a) The participant usually communicates through the shown non automatic system. (b)  The participant selects letters using Ticker. The shown poster
was the only visual assistance he received to help him remember the clips. (c) A close-up of the poster in (b). (d) The participant's initial
reaction to Ticker in 5-channel mode.
 }
  \label{fig:case_study_pictures}
\end{figure*}

Photos from the first session of the case study described are provided in \figref{fig:case_study_pictures}.
The case study was done in collaboration with Special Effect~\cite{specialeffect}, a charity based in Oxford. 
They were present at all sessions, along with the participant's carer. The staff from Special Effect provided invaluable advice, and also access and visits to other potential users who 
are not mentioned in this paper; time allowed for only one participant to undergo a full evaluation.
The photos were also  kindly provided by Special Effect. 

Despite the participants initial doubt (depicted in \figref{fig:case_study_pictures}(d)) he was able, to his own surprise, to select letters easily using Ticker in 5-channel mode. He communicated that he had fun doing the user trials. 

\section{A Note on Brain-Computer Interfaces (BCI)}
\label{subsec:BCI}

In this paper we have focussed on single-switch text entry methods. However, we believe Ticker's possible  resilience to noise could potentially make it an ideal candidate for Brain-Computer Interfaces (BCIs), which translate brain activity into computer actions (e.g.~\cite{Hoffman2007_2,Millan2010, Tan2010}). Due to the noise, text entry rates in BCIs are extremely low \cite{Millan2010, Blankertz2007, Welton2014, Wills2006}. The main contributing factors to such resilience are its predictive texting traits, the customisable  language model  and its robustness to long click-time delays and false positives. Users can initially make use of the 1-channel configuration, and gradually progress to more channels if applicable. One might have to repeat the alphabet more than once, as mentioned in \secref{sec:audio_seq}.

Brain-Computer Interfaces (BCIs) translate brain activity into computer actions (e.g.~\cite{Hoffman2007_2,Millan2010, Tan2010}). To convert brain activity into a signal that can be reliably used to control a switch for a scanning system is difficult as the signal-to-noise ratio is generally low.
 Several methods have been  developed to capture brain activity each with its pros and cons; see \cite{Tan2010} for a summary. 
 Two well-known techniques to capture brain-activities are Electroencephelography (EEG) and Functional Magnetic Resonance Imaging (fMRI).

 EEG is the predominant  technology, 
 where electrodes are placed on the head to measure weak electrical potentials~\cite{Tan2010}. The technique has low spatial resolution (\nu{2-3}{cm} at best) and requires careful setup.
The latency 
is low (tens of milliseconds). On the other hand,  fMRI has high spatial resolution ($<$ (\nu{1}{cm}), but high latency ((\nu{5-8}{seconds}). 
 EEG can therefore capture a  (non-specific) brain signal quickly, whereas fMRI can capture the user's thoughts with much higher accuracy albeit at a slower rate. 
 Tan and Nijholt~\cite{Tan2010} mention  that (\nu{2}{cm} on the cerebral cortex could make it difficult to distinguish if the user is listening to music or conducting a hand motion. Text-entry methods controlled by EEG should therefore be highly resilient to false positives, whereas if controlled by fMRI, they should be highly resilient to long delays. 

 Due to the noise, text-entry rates are extremely low and typically measured in characters per minute~\cite{Millan2010}. Blankertz~\cite{Blankertz2007} reports a text-entry rate of \nu{7.6}{\nu{char}{\nu{/}{min}}}, controlling Hex-o-Spell with two switches (only two subjects were tested).  Millan et al.~\cite{Millan2010} mention Hex-o-spell in the context of state-of-the art BCI-spelling devices (in 2010), and as an improvement on the Thought-Translation-Device (with a reported text-entry rate of \nu{ 0.5}{\nu{char}{\nu{/}{min}}}).

 In a recent study (2014),  Welton et al.~\cite{Welton2014} reports on the use of Dasher in a BCI context. The pioneering study by Wills and MacKay~\cite{Wills2006}
 thought it a viable text-entry method in this context due to its personalised language model and the ability to navigate towards a symbol instead of selecting one symbol at a time (making it more resilient to the noisy EEG data). However, there was uncertainty regarding the cognitive load of this visually intensive task. Welton et al.~\cite{Welton2014}
 tested seven users with a wide range of disabilities. They found that Dasher-BCI was not the answer for all the users, but it may be viable in some cases,  and justifies more extensive testing. For example, one  user with cerebral palsy who was unable to use the QWERTY keyboard or Dasher-Mouse, could use Dasher-BCI, typing at \nu{4.7}{\nu{char}{\nu{/}{min}}}.

 \section{A Note on Error Corrections in Ticker}

 Error corrections in Ticker are used only in extreme circumstances, as noise compensation allows for a large variety of implicit error correction.

 In some cases, two words can strongly compete against each other, especially if the user clicked inaccurately and the intentional word is short. A typical example will be ``in\_'' and ``is\_''. In the 5-channel mode ``n'' and ``s'' are nearest neighbours (see \figref{fig:letter_config}(a)). If the user clicked slightly inaccurately while aiming for one of these two letters, the probability would typically be split close to equal between the two words to say 0.45 and 0.5. The word would then typically have to be repeated until the confusing letter is reached again, at which point the user would typically resolve  his/her previous error.   

 This aforementioned problem can, of course, be improved by not allowing the letters ``n'' and ``s'' to be neighbours in the first place. A second way to work around this problem would be to change the word-selection heuristic to evaluate the sum of the top three posterior word probabilities instead. The user can then perhaps select between the top three words in some way. This would be slightly complicated, since care has to be taken not to break the user's thought process, in case he/she has to resume with letter selections if the intentional word is not in the top three.  
It should be noted, however, that during all user trials and simulations it was found to be extremely rare for the intended word not to end up in the top three words, especially by the time the system fails (which is a rare event in itself). 


\bibliographystyle{IEEEtran}
\bibliography{references}

\end{document}